\documentclass[conference, letterpaper]{IEEEtran}
\pdfoutput=1
\usepackage{cite}
\usepackage{hyperref}

\hypersetup{
  pdfinfo={
    Title={Strategies for Using Proximal Policy Optimization in Mobile Puzzle Games},
    Author={Jeppe Theiss Kristensen}
  }
}

\usepackage[utf8]{inputenc}
\usepackage{listings}
\usepackage{rotating}
\usepackage{siunitx}
\usepackage{amsmath,amssymb,amsfonts}
\usepackage{algorithmic}
\usepackage{graphicx}
\usepackage{textcomp}
\usepackage{xcolor}

\def\BibTeX{{\rm B\kern-.05em{\sc i\kern-.025em b}\kern-.08em
    T\kern-.1667em\lower.7ex\hbox{E}\kern-.125emX}}

\begin{document}

\title{Strategies for Using Proximal Policy Optimization in Mobile Puzzle Games}

\author{\IEEEauthorblockN{Jeppe Theiss Kristensen}
\IEEEauthorblockA{ 
\textit{IT University of Copenhagen/Tactile Games}\\
Copenhagen, Denmark \\
jetk@itu.dk}
\and
\IEEEauthorblockN{Paolo Burelli}
\IEEEauthorblockA{
\textit{IT University of Copenhagen/Tactile Games}\\
Copenhagen, Denmark \\
pabu@itu.dk}
}

\maketitle

\begin{abstract}

While traditionally a labour intensive task, the testing of game content is progressively becoming more automated. 
Among the many directions in which this automation is taking shape, automatic play-testing is one of the most promising thanks also to advancements of many supervised and reinforcement learning (RL) algorithms. 
However these type of algorithms, while extremely powerful, often suffer in production environments due to issues with reliability and transparency in their training and usage.

In this research work we are investigating and evaluating strategies to apply the popular RL method Proximal Policy Optimization (PPO) in a casual mobile puzzle game with a specific focus on improving its reliability in training and generalization during game playing.

We have implemented and tested a number of different strategies against a real-world mobile puzzle game (Lily's Garden from Tactile Games). 
We isolated the conditions that lead to a failure in either training or generalization during testing and we identified a few strategies to ensure a more stable behaviour of the algorithm in this game genre.

\end{abstract}


\section{Introduction}

Human game testing is an expensive and slow process.
It usually requires the full attention of the testers, and there are limitations on how fast humans can operate.
Game developers are therefore increasingly starting to use automated play testing.
However, developing and implementing such methods in practice has its problems -- the methods tend to require a very specific setup for one game, and trying to adapt it to other environments may sometimes break the algorithm and render it useless.
In this paper we therefore set out to explore both novel and common strategies for ensuring a stable implementation of a reinforcement learning (RL) play-testing agent in a mobile puzzle game in a production setting.

\begin{figure}[t]
    \centering
    \includegraphics[width=0.95\columnwidth]{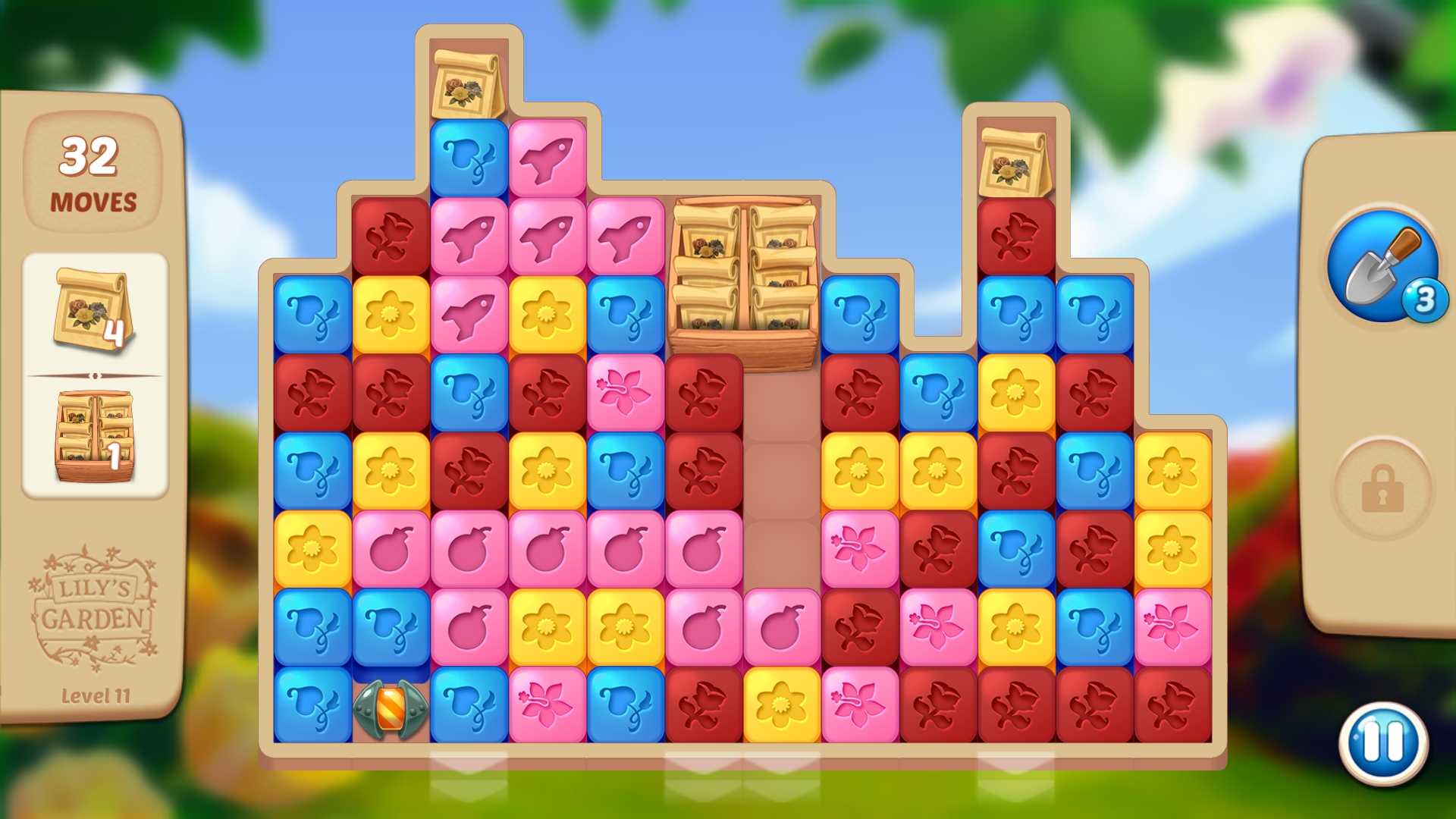}
    \caption{Level 11 from Lily's Garden. The left hand side shows number of moves left to finish the level, and the board pieces below indicate which and how many pieces to collect before completing the level. Collecting the objectives is done by clearing them off the board, which can be done by clicking on two or more basic pieces of the same color, or using power pieces that clear an entire row/area. Power pieces can be created by matching 5 or more basic pieces.}
    \label{fig:LG level 11}
\end{figure}
A popular choice for creating play testing tools is reinforcement learning, and research in this field is moving fast.
Novel algorithms and updates to current state-of-the-art methods are constantly being introduced in the latest publications, showing better performance on typical frameworks such as the Arcade Learning Environment \cite{Bellemare2015TheAgents}.

However, contrary to these kind of one-shot evaluations, adapting these methods in a production environment in a company requires additional considerations -- such as ease-of-use and long-term reliability.
Unlike these benchmark games, production games are updated frequently, and it can not be expected to be possible to draw on expert knowledge at any time in case something goes wrong.
Until more focus has been put on strategies on \textit{how} to use these methods, adoption of these methods in the industry will be slow at best.

In this research work we focus on the challenges of implementing the popular RL method Proximal Policy Optimization (PPO) \cite{Schulman2017}, a widely used algorithm available in various RL libraries (OpenAI Baselines/stable-baselines \cite{baselines, stable-baselines}, TF-Agents \cite{TFAgents}, Unity ML-Agents \cite{Juliani2018}), in a mobile puzzle game called Lily's Garden by Tactile Games (Fig. \ref{fig:LG level 11}).
While other RL methods may also work in this environment, we choose to only focus on PPO since it is one of the two main algorithms implemented in Unity ML-Agents and thus widely accessible to game developers that use Unity.

Our contribution is two-fold:
\begin{itemize}
    \item We explore different setups for training an agent in a mobile puzzle game and determine a set of hyperparameters and setups that enable the agent to some extend play both seen and unseen levels competently.
    \item We highlight that the impact of some PPO variations are not fully understood and can easily lead to unexpected learning behaviours. We then suggest strategies for avoiding such behaviours and ensure a more stable training.
\end{itemize}

This paper is structured as follows:
First we introduce the game environment that we will use for testing. Next we present the basics of the PPO algorithm and discuss the specific implementation we use. This is followed by the experiments section where we present the various setups we tested and highlight the main difficulties and problems encountered during training of the agent. Lastly we discuss which methods and strategies that are feasible to employ in a production setting and identify areas that need improvement.

\section{Related work}

When it comes to creating agents for playing games, reinforcement learning (RL) and deep learning methods have started to become a staple and have been used to play a large variety of games, ranging from arcade games to first-person shooter games \cite{Justesen2019DeepPlaying}.

Each genre has its own challenges, and some approaches work better than others in different settings.
It is therefore relevant to consider which approaches that have been used for play-testing in similar game genres.

In Atari games, some of the state-of-the-art approaches using pixel data or memory-features as input are deep Q-learning (DQN) \cite{Mnih2015Human-levelLearning, Mnih2013} and variations thereof (such as Rainbow \cite{Hessel2018Rainbow:Learning}), and actor-critic approaches like PPO \cite{Schulman2017} and soft actor-critic \cite{Haarnoja2018} in \cite{Christodoulou2019SoftSettings}.

The \textit{MuZero} algorithm introduced by Schrittwieser et al. ~\cite{Schrittwieser2019MasteringModel} uses a combination of tree-search planning and a learned model of the environment and is capable of playing Go, Chess, Shogi and the Atari games.
However, how to deal with stochastic transitions was not examined.

As for approaches used specifically on puzzle games, other approaches have also been directly applied.
Gudmundsson et al. ~\cite{Gudmundsson2017} treat the task as a classification problem and train a convolutional neural network on player data. 
Their method beats state-of-the-art Monte-Carlo Tree Search algorithms in terms of difficulty prediction and training time and has been used actively for a year by the time of publication.
However, this method requires play-through data which may not always be available.
Mugrai et al. ~\cite{Mugrai2019AutomatedGames} use a MCTS method with an evolutionary strategy where the fitness function is used to mimic specialised player personas/strategies with different goals, such as maximising score or minimising moves used.
This aspect of creating human-like agents is indeed important if they are to be used as a play-testing tool, which is also highlighted by Zhao et al. ~\cite{Zhao2019} where the agents are evaluated by considering both skill and style.
A comparison of three popular methods (DQN, PPO and A3C) in a custom match-3 game is done by Kamaldinov et al. ~\cite{Kamaldinov2019DeepGame} which shows that the A3C method achieves the highest accumulative reward while the PPO and DQN methods perform worse than random.
They use a custom match-3 environment, though, so it is not clear if these results reflect real-world results in puzzle games.
An example of training an agent using actual games levels can be seen in the Unity blogpost \cite{TrainingBlogb}, where an agent for playing Snoopy Pop using ML-Agents in \cite{TrainingBlogb} is attempted using a actor-critic method (SAC, \cite{Haarnoja2018}) and imitation learning (GAIL, \cite{Ho2016GenerativeLearning}).
Although a slightly different genre, it shows that the training can be sped up using sample efficient methods and a player to guide the agent initially.
However, a similar approach is not efficient in games like Lily's Garden since in those cases it is not necessary to simulate physics.
Furthermore, there are more than 1500 levels available so deciding which levels to train on or alternatively have a player play through all of them is not scalable.

When it comes to using such automated systems in a production setting, reliability and accessibility of the algorithm are critical components.
The less interference required, the better, and when something goes wrong, identifying the points of failure easily is important so it can be fixed quickly and not waste resources.
RL systems, especially PPO approaches, tend to be the antithesis of these requirements: they tend to be brittle \cite{Hamalainen2018}, and the stability tends to be implementation-dependant \cite{Ilyas2018}.
It is therefore important to consider not just the algorithms but also the strategies of how a play-testing tool should be developed.

Such a tool also needs to be able to generalise to new levels, and one problem that appears in many RL papers is overfitting to an environment \cite{Zhang2018}.
Ways to diagnose and improve the generalisation in deep RL systems have been examined is various works \cite{Packer2018AssessingLearning, Zhang2018}.
Farebrother et al. ~\cite{Farebrother2018GeneralizationDQN} find that dropout and $\ell 2$ regularisation with a DQN method improve generalisation.
This is also supported by the findings by Cobbe et al. ~\cite{Cobbe2018} where data augmentation, batch normalisation and stochasticity were also found to improve generalisation in an implementation of PPO.
Adding entropy regularisation also helps find smoother solutions but is very environment dependent \cite{Ahmed2018}.
Variations in the levels by using procedural content generation methods can also improve generalisation and help learn more difficult levels \cite{Justesen2018IlluminatingGeneration}.
Avoiding undesirable and dangerous actions may also help the agent learn more efficiently because of better and safer exploration strategies \cite{Zahavy2018LearnLearning}.
Having the system learn which actions to eliminate has been the focus in some recent works \cite{Seurin2019, Zahavy2018LearnLearning, Alshiekh2017}. 
In addition to learning action blocking, Kenton et al. ~\cite{Kenton2019} also use an ensemble model in both a DQN and PPO setup.
While the DQN method showed improvements, the PPO experiments showed little improvement compared to the baseline.

\section{Environment}

\begin{figure*}[th]
    \centering
    \includegraphics[trim=0 10 0 50, clip,  width=0.90\columnwidth]{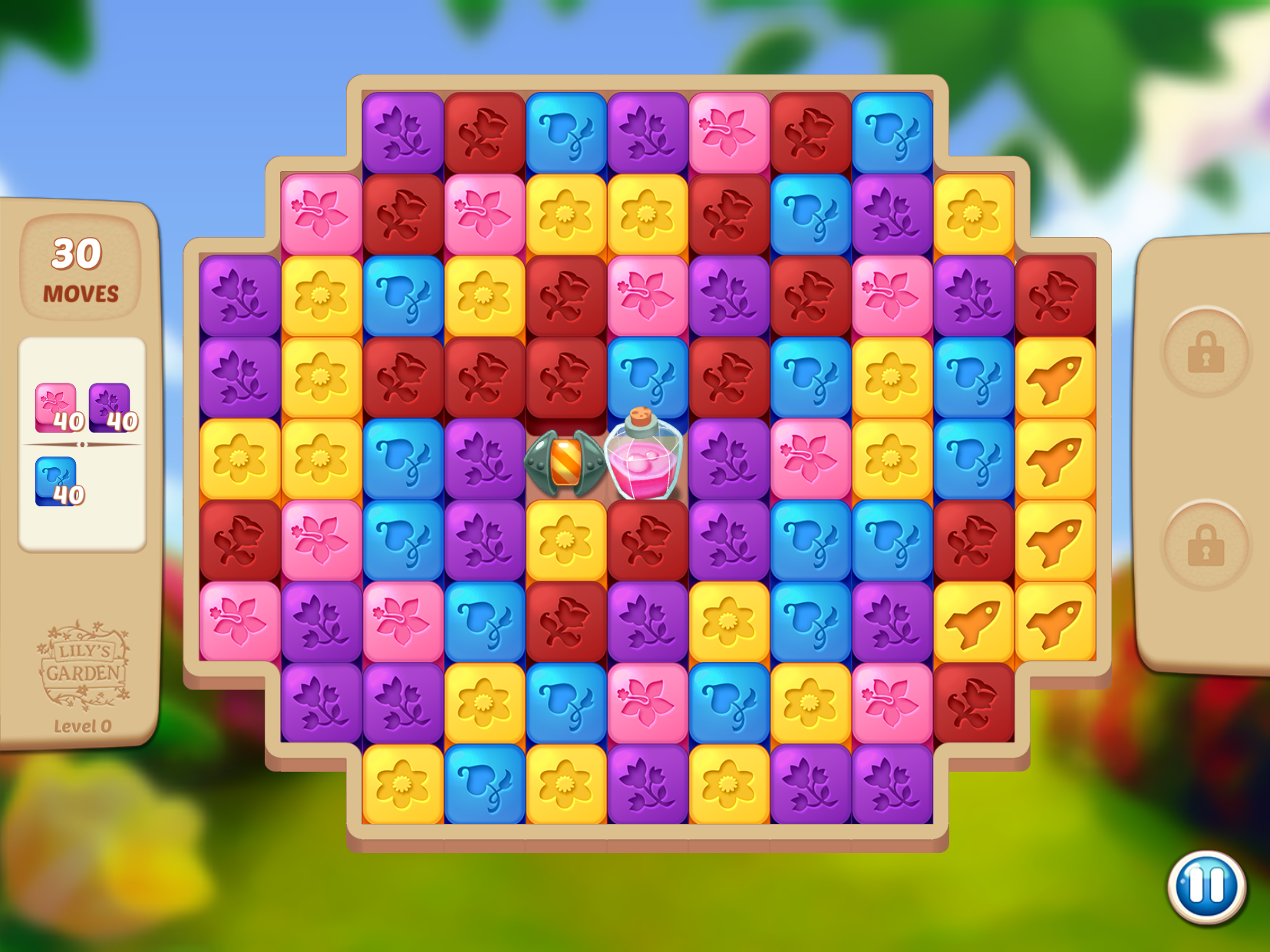}
    \includegraphics[width=0.95\columnwidth]{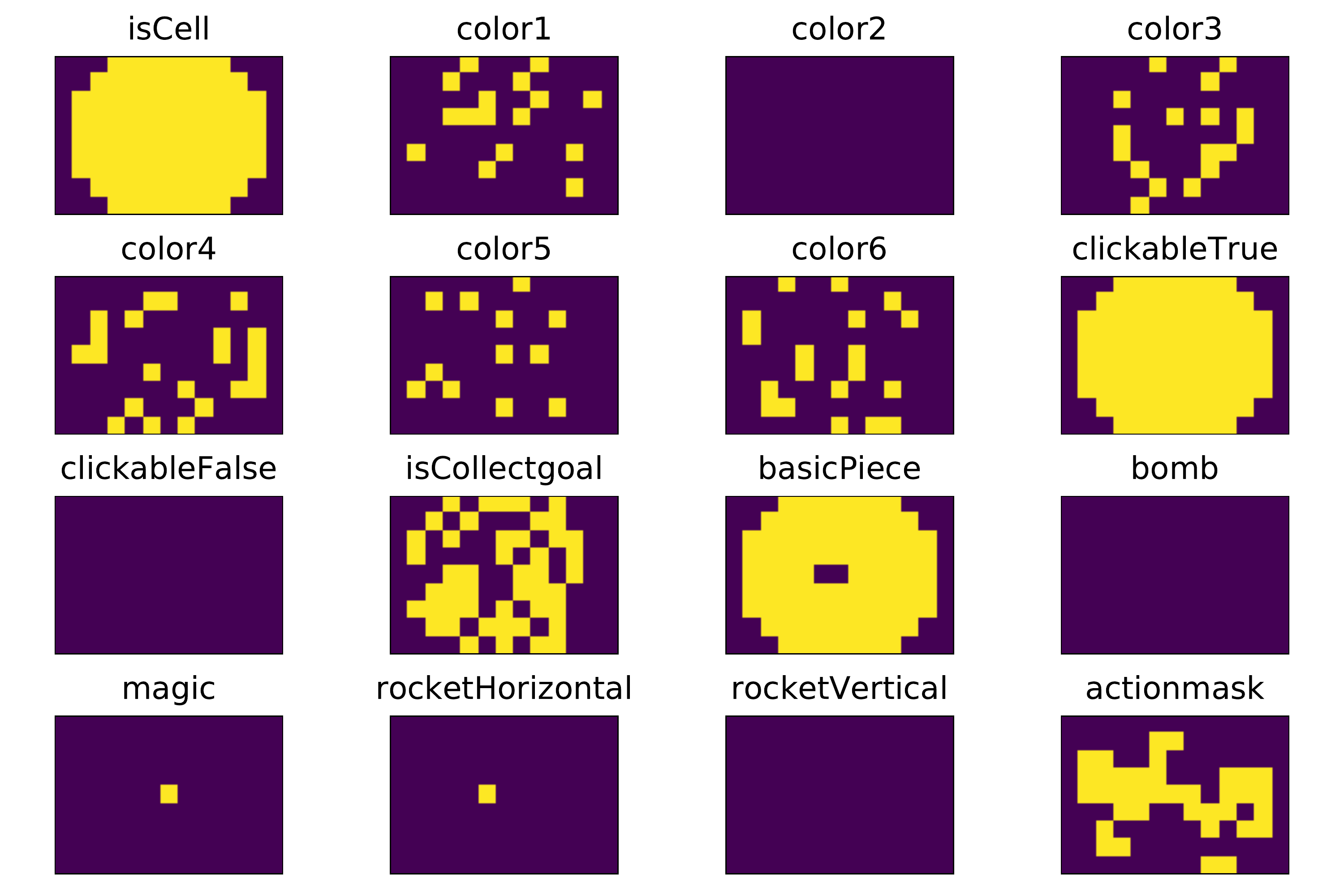}
    \caption{Example of how an in-game level looks like and how the game board is represented using different channels corresponding to certain board piece attributes. Note that the last channel, the action mask, is only included in certain experiments.}
    \label{fig:level representation}
\end{figure*}

In this paper we focus on one game, Lily's Garden\footnote{\href{https://play.google.com/store/apps/details?id=dk.tactile.lilysgarden}{Android}, \href{https://apps.apple.com/us/app/lilys-garden-design-relax/id1437783446}{IOS}}.
It is a free to play casual puzzle mobile game where you progress through the main story by completing levels.
The main gameplay is matching similar colored pieces and thereby collecting objectives (collectgoals), which must be done before running out of moves.
The game board has a maximum size of 13 by 9, and in each position, board pieces with various attributes may be placed.
The basic pieces can be destroyed/collected if two or more of the same color are next to each other and will create power pieces if 5 or more are next to each other.
The power pieces can be clicked at any time and destroy everything in for example a line or circle around the position.
Lastly there are unclickable board pieces, or blockers, that can be removed by matching basic pieces next to it or sometimes only by using a power piece.
An example of a level is shown in Fig. \ref{fig:LG level 11}.

We set up an OpenAI gym environment that connects to headless version of the game (no graphical interface) which, for speed purposes, allows us to play through levels without rendering any graphics.
We define a rich reward function where the reward is calculated at each step as: $r = c_{\textup{collection}} n + c_{\textup{completion}} - 0.1$, where $c_{\textup{collection}} = 0.05$, $n$ is the number of collected collectgoals and $c_{\textup{completion}} = 1$ if all collectgoals have been collected. The negative term, $-0.1$, is added to encourage the agent finishing the level faster as to not get a large negative accumulated reward. Given that a typical level has around 50 collectables and requires up to 25 moves to complete, the expected final reward is $R \approx 50 \cdot 0.05 - 25 \cdot 0.1 + 1 = 1$ (not considering discount).

Since each board piece may be of the same type (e.g. basic or blocker) but different attributes (e.g. color or gravity), one-hot encoding each board piece by the unique combination of attributes may lead to a very large and sparse representation, as seen in \cite{Gudmundsson2017}.
Instead we choose to represent the observation space by using layers that correspond to the attributes of all the board pieces in a given position (see Fig. \ref{fig:level representation}).
Specifically, we represent the following attributes with a layer giving a total of 15 channels:
\begin{itemize}
    \item \textsc{isCell}: used to define shape of game board
    \item \textsc{color}: one-hot encoding of 6 unique colors
    \item \textsc{isCollectgoal}: if board piece is a collectgoal
    \item \textsc{isClickable[True/False]}: one layer for clickable, another for non-clickable since a non-clickable piece may be on top of another
    \item \textsc{id}: one-hot encoding of \textsc{basicPiece}, \textsc{rocketHorizontal}, \textsc{rocketVertical}, \textsc{bomb} and \textsc{magic}
\end{itemize}
This approach also has an advantage when it comes to generalizability for future versions because the observation space will not depend on graphics updates and new types of board pieces are typically made up of a combination of existing attributes.
The action space consists of $9 \times 13 = 117$ discrete actions, corresponding to each square of the game board.

\section{Methods}

The typical reinforcement learning problem consists of an agent that interacts with an environment and receives a reward depending on the action.
This loop may then continue indefinitely or until the episode ends.
The main purpose of the algorithm is then to learn a behaviour that maximises the accumulated reward \cite{BartoSutton}. 


In the original form, PPO refers to a family of policy gradient methods that optimize a (clipped) surrogate objective function using multiple minibatch updates per data sample. However, the exact implementations in various libraries may be slightly different because of other additions such as value scaling or batch normalisation \cite{Ilyas2018}. Common for them all is the suggested function to optimise, which is the sum of several loss functions and is given by

\begin{equation}
    L^{CLIP+VF+S}(\theta)_t = \hat{\mathbb{E}}_t \left[ L^{CLIP}(\theta) - c_1 L_t^{VF}(\theta) + c_2 S[\pi_\theta] (\theta) \right],
    \label{eg:loss function all}
\end{equation}
where $L^{CLIP}(\theta)$ is the clipped surrogate objective function, $L_t^{VF}(\theta)$ is the value function squared-error loss, $S$ is an entropy bonus and $c_1$ and $c_2$ are coefficients. The $L^{CLIP}(\theta)$ term ensures that the policy updates will not be too large, and the $L_t^{VF}(\theta)$ term is to ensure that the loss from both policy and value functions of the neural networks are accounted for. The $S$ entropy term encourages a more random policy (i.e. more exploration) so a larger entropy coefficient $c_2$ will encourage more exploration.


\subsection{Implementation}
\label{sec:methods: action masking}

Since we want to investigate strategies for implementing PPO in a production environment, we choose to go with a widely used code library.
Some of the notable RL libraries are OpenAI Baselines\footnote{\url{https://github.com/openai/baselines}} and Unity ML-Agents\footnote{\url{https://github.com/Unity-Technologies/ml-agents}}.
For the following experiments we choose to use stable-baselines, which is a fork from OpenAI Baselines but follows the same algorithmic implementation of PPO.

We test out three different strategies which will be described in this section.
These strategies are:
\begin{itemize}
    \item Color shuffling (CS)
    \item Resetting
    \item Action mask
\end{itemize}

\textit{\textbf{Color shuffling}} refers to swapping the color channels in the observations randomly.
While color shuffling is done in the post-training evaluation for all models to simulate how levels are designed, we want to test how effective it is to also include this strategy during training.
It should also help prevent overfitting -- even though it is random which board pieces that drop down and replace cleared pieces, the initial setup are usually predetermined (see Fig. \ref{fig:levelcollage}) which may lead to strong overfitting.

\textit{\textbf{Resetting the environment}} commonly happens at the end of episodic environments, which in this case could be when the level is completed or failed.
However, the level move limit is subject to change because of design considerations, and we already add a penalty at each step to encourage it finishing faster.
Imposing a move limit does therefore not make much sense.
What we do try with the reset strategy, though, is imposing a total step limit, which includes both valid and invalid moves.
The reasoning behind this strategy, similar what is given in \cite{Liu2019} using restarts in Angry Birds, is that the agent is prevented from exploring useless states that it will not learn anything from.

Before deciding what the maximum episode length could be, two things should be considered. 
One is how the typical PPO implementation samples observations.
In our case, we sample 256 observations before training on these minibatches.
This means that if we reset after 256 total steps, we may end up with a full minibatch of bad training samples, which is undesirable.
Secondly the typical steps required to pass a level is generally around 50.
Levels that require more steps are rare since it would be very frustrating for players to almost finish a level but ultimately fail after, say, 100 steps rather than 50.
We therefore choose to reset after 100 steps which should ensure at least some good observations and still allow the agent to complete a level.

\textit{\textbf{Using an action mask}} during training is the last strategy we explore.
While we give a penalty for selecting an invalid action, preventing the agents from selecting certain catastrophic or invalid actions may lead to more efficient learning.
The question is how this limitation should be implemented.
We use two different approaches for creating action masks in the following experiments -- a hard and a soft action mask.

With the hard action mask, the invalid actions are completely masked when sampling from the policy distribution. 
In practice, this is done by adding the mask to the logits of policy distribution, where valid actions have a value of $0$ and invalid action a value of $-\infty$. 
This is slightly different than in the ML-Agents library where a small probability $\epsilon$ is added to the action probabilities which prevents $\infty$ values but also allow invalid actions to be taken, albeit with a very low probability.
The way sampling is done in the stable-baselines library is by using a Gumbel-max trick.\footnote{\url{https://github.com/hill-a/stable-baselines/blob/a57c80e0636582995d602309d2ea5547c0d58e61/stable_baselines/common/distributions.py\#L323}}
Specifically, noise following a Gumbel distribution (computed by taking the negative logarithm twice of uniformly distributed noise) is added to the logits which ensures the sampling will follow the underlying probabilities of the actions.

The soft action mask is a kind of forward model of the environment. 
Specifically we add the action mask to the observation space as an additional channel, as illustrated in the last panel in Fig. \ref{fig:level representation}.
The reason for calling this a soft action mask is because it does not directly prevent invalid actions from being taken although it might significantly reduce the probability.
The soft action mask model is denoted with V2.

Since the game simulator does not provide a method for getting the action mask, we define it ourselves.
It follows the basic rules that an action is valid if at least two \textsc{basicPieces} are adjacent and of the same color, or if there is a power piece in the cell.
While this is not true for later levels, it is sufficient for the first 11 levels that we test on.

Lastly, we also want to evaluate if it makes a difference to continue training after the learning curves have plateaued since shorter training times allow for quicker iterations and thus easier testing in a production environment.
These long-trained models are denoted in the post-training evaluation figures with (late).

\section{Experiments}

\begin{table}[b]
    \centering
        \caption{Overview of models and used entropy coefficient (EC)  as well as which training step checkpoint used for post-training evaluation. The section in which the results of said models are also shown. CS: color shuffle.}
    \label{tab:post-training evaluation models}
\begin{tabular}{l|SSc}
\multicolumn{1}{c}{\textbf{Model}} & \multicolumn{1}{c}{\textbf{EC}} & \multicolumn{1}{c}{\textbf{Step} ($\times 10^6$)} & \textbf{Section} \\ [0.5ex]
\hline \\ [-2ex]
Baseline & 0 & 0.35 &  \ref{sec:generalisation}\\
Baseline & 0.001 & 0.20 & \ref{sec:baseline}\\
Baseline & 0.01 & 11 & \ref{sec:generalisation}\\
CS & 0.001 & 0.20 & \ref{sec:generalisation}\\
CS & 0.01 & 14 & \ref{sec:generalisation}\\
CS+reset & 0.01 & 0.35 & \ref{sec:max episode length} \\
CS+reset+mask & 0.01 & 6.5 & \ref{sec:action mask} \\
CS+reset+maskV2 & 0.01 & 4.0 & \ref{sec:action mask} \\
CS+reset (late)  & 0.01 & 14 & \ref{sec:max episode length} \\
CS+reset+maskV2 (late) & 0.01 & 14 & \ref{sec:action mask}
\end{tabular}
\end{table}

\begin{figure}[t]
    \centering
    \includegraphics[width=0.95\columnwidth]{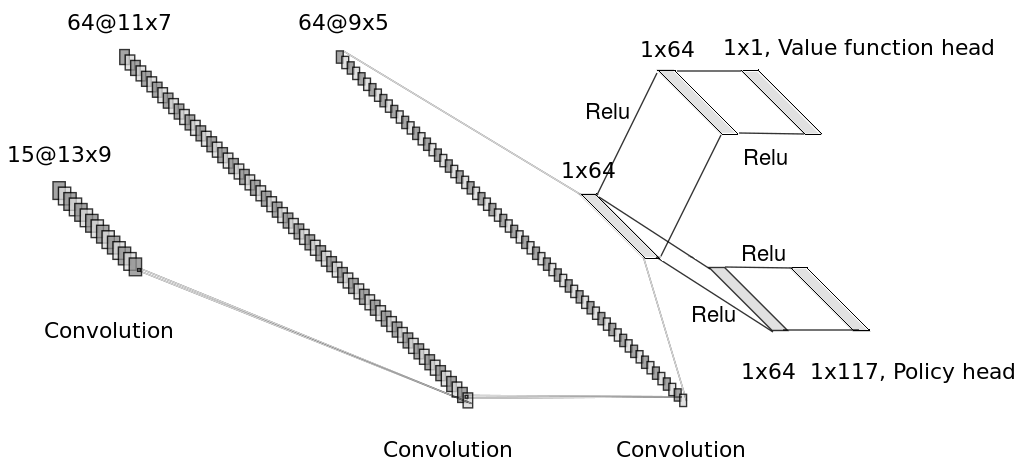}
    \caption{The network architecture of the agent.}
    \label{fig:CNN policy}
\end{figure}

\begin{figure}[t]
    \centering
    \includegraphics[width=0.90\columnwidth]{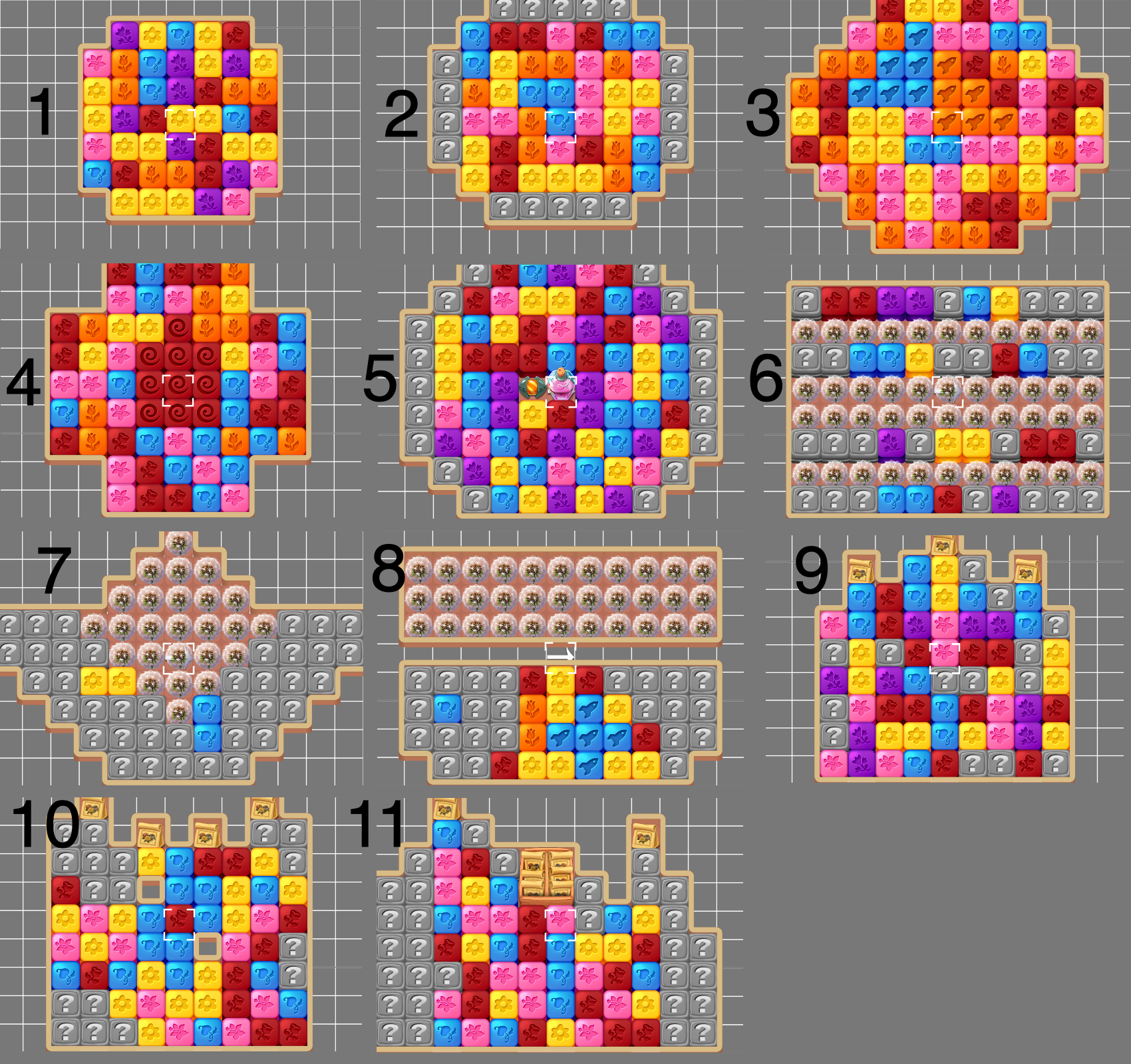}
    \caption{Levels used for the experiment. Board pieces with question marks are assigned a random color on level start, while every other board piece is hardcoded.}
    \label{fig:levelcollage}
\end{figure}

\begin{figure*}[ht]
    \includegraphics[width=1.95\columnwidth]{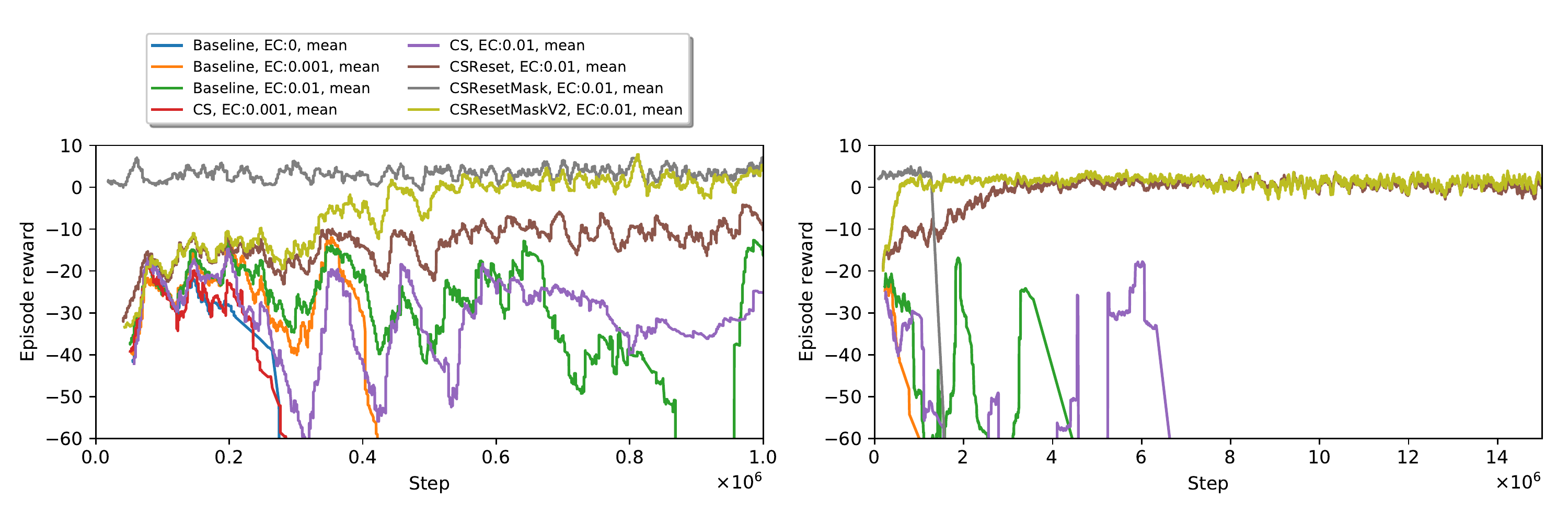}
    \caption{Learning curves for the tested model. The left figure shows a zoomed in version on the first 1 million steps. CS refers to models trained with color shuffling.}
    \label{fig:learning curves}
\end{figure*}

\begin{figure*}[t]
    \centering
    \includegraphics[width=1.95\columnwidth]{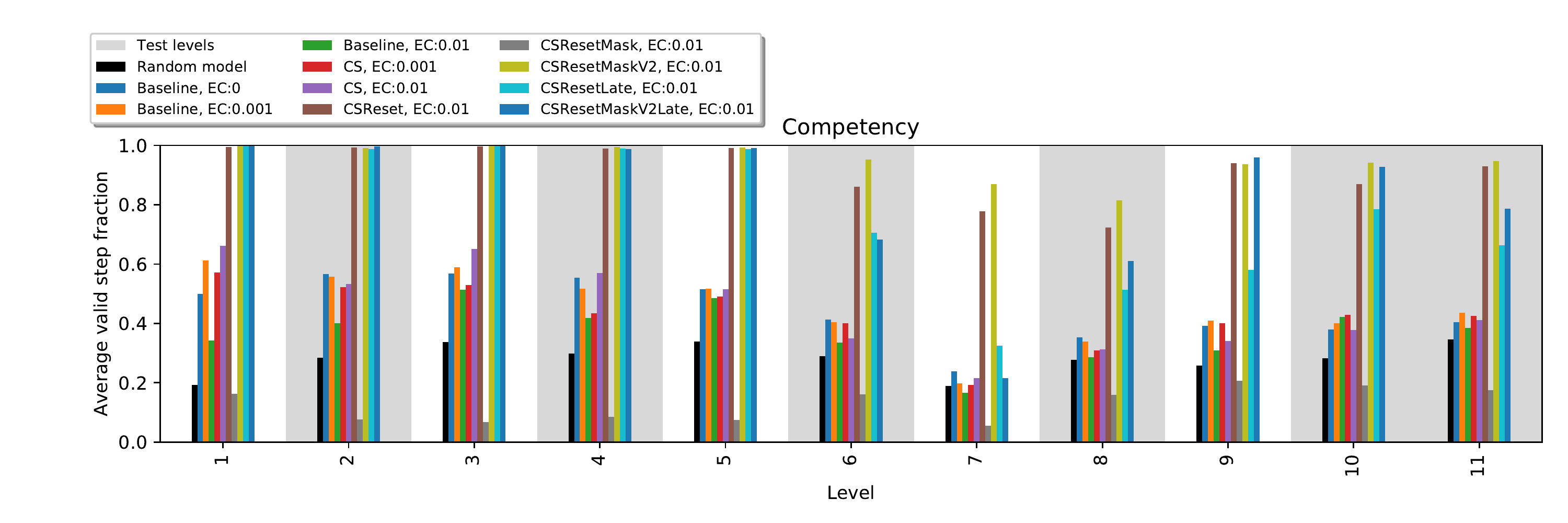}
    \caption{Number of valid moves per level for each model. CS refers to models trained with color shuffling. The grey shaded levels are the unseen test levels}
    \label{fig:valid moves percentage}
\end{figure*}

\begin{figure*}[t]
    \centering
    \includegraphics[width=1.95\columnwidth]{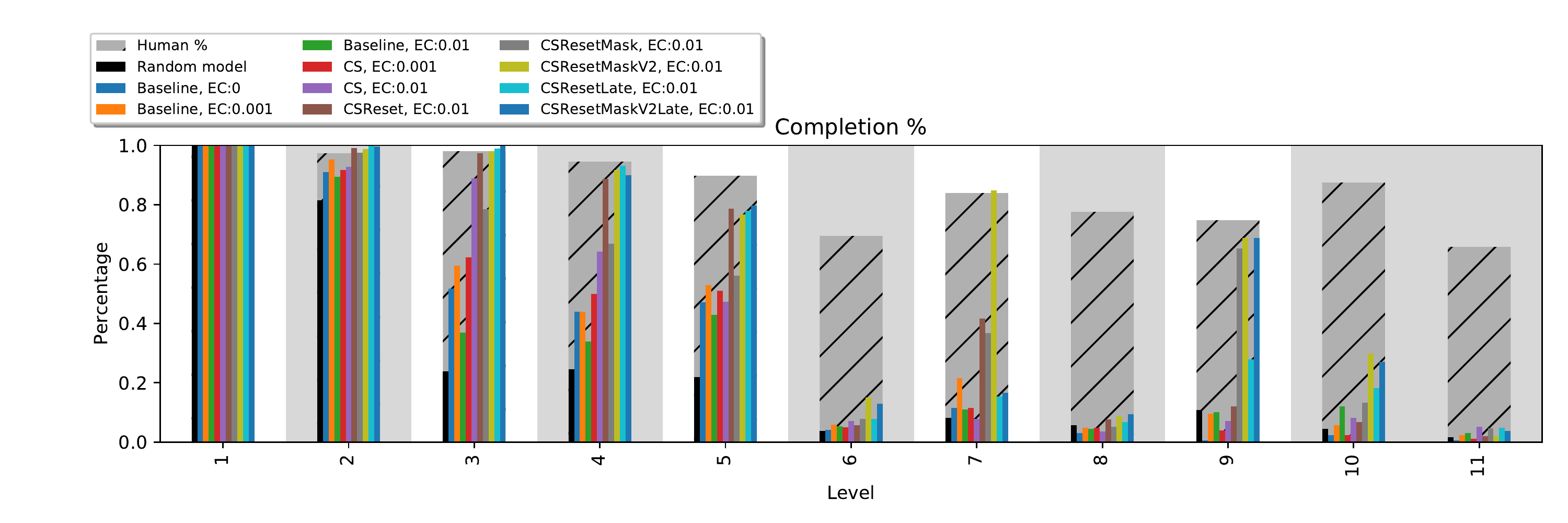}
    \caption{Completion rate for each model given the level limit. CS refers to models trained with color shuffling. The unseen test levels are shown with a grey shaded area as in \ref{fig:valid moves percentage}, while the human completion rate is shown as the large grey hatched bar.}
    \label{fig:completion_pct_article}
\end{figure*}

We carried out a number of experiments to test the performance of the PPO algorithm in our environment. We used the PPO2 implementation from the Python RL library stable-baselines \cite{stable-baselines} and a custom CNN policy (Fig. \ref{fig:CNN policy}). 

Each of the experiments are evaluated similarly to \cite{Cobbe2018} where the trained agent is tested on unseen levels in order to evaluate its ability to generalize.
This is done by training on 5 chosen levels (1, 3, 5, 7 and 9) selected randomly and uniformly during training and validated using an additional 6 levels (2, 4, 6, 8, 10 and 11). 
With the exception of level 11, these levels include two unique blockers, and splitting the levels accordingly ensures both that the tutorial levels and trained on and both the training and test sets will include at least one level containing any of the blockers.
Level 11 has a third unique blocker so we include that level in the evaluation to see how the agent performs with completely unseen mechanics.
An overview of the levels is shown in Fig. \ref{fig:levelcollage}.

\subsection{Evaluation Metrics}

During training we consider the accumulated reward/learning curve as the evaluation statistic.
For the post-training evaluation we do not want to only estimate if the agent can finish the level within the actual in-game max moves but also how competent it is compared to a random agent.
We therefore allow up to 2000 total steps and do not use an action mask.
We also shuffle the colors during evaluation for all models in order to simulate actual in-use performance, since the different colors of the board pieces only affect the aesthetics of the game and are used interchangeably.

We will consider two post-training evaluation metrics:

\textbf{\textit{Competence}}
is the reciprocal average index (starting with 1) of the first valid action after sampling actions using the action probabilities without replacement. 
Taking the reciprocal value corresponds to estimating the average valid step percentage and can be thought of as a proxy for how well the agent understand the basic match-2 mechanic of the game.

\textbf{\textit{Level completion percentage}}
is calculated by imposing the level move limit on the agent.
We also include actual player data.
It should be noted that the player completion percentage is estimated by taking the number of level completions over total number of level attempts.
However, the level attempts include successes, failures \textit{and} abandoning the game, where the latter may happen if the game for example crashes, other technical failures or simply just giving up on a level.
Abandoning the game typically happens less than 5\% of the time, though, so this effect should be minor.

\subsection{Model Setup}

We did a preliminary analysis training various models with different hyperparameters to find a stable configuration. 
While we also experimented with reward shaping and state representations, the key changes required to get the PPO algorithm to work with Lily's Garden was changing the minibatch size, number of steps per update and number of actors. 
We found that setting nminibatches to 64 (default: 4), n\_steps to 256 (default: 128) and the number of parallel actors is set to 8 gave a good balance between speed and stability of the algorithm. This is not unexpected as these changes from default ensure a smoother gradient and faster and more stable training \cite{Heess2017EmergenceEnvironments} and thus more stable training.

We use a custom neural network setup as shown in Fig. \ref{fig:CNN policy}.
It uses three 2x2 convolutions with filter size 64 and leaky relu activations, which are fed into two fully connected 64 layer for the actor and critic heads respectively.

The above hyperparameters are kept the same throughout the experiments except for the entropy coefficient, which will be discussed further in Section \ref{sec:baseline} and \ref{sec:generalisation}.
That setup will serve as a baseline model where no special strategies for training are used.
For the other experiments, we use the three aforementioned strategies in Section \ref{sec:methods: action masking}.

\section{Results}
\label{sec:baseline}

The learning curves for every model can be seen in Fig. \ref{fig:learning curves}, the valid move percentages in Fig. \ref{fig:valid moves percentage} and the completion rate in Fig. \ref{fig:completion_pct_article}. Table \ref{tab:post-training evaluation models} shows at which step each model was evaluated as well as in which of the following sections they are discussed further.



In this section, we only consider the baseline model with an entropy coefficient (EC) of 0.001.
The other two baseline models are discussed in the next section.

Looking at the learning curve of the Baseline EC: 0.001 model in Fig. \ref{fig:learning curves}, it can be seen that the agent quickly learns as reflected in the increase in episode rewards.
However, after $~400.000$ steps, the episode reward sharply decreases and completely breaks the training.
The same behaviour was also observed in other experiments during the initial analysis.
This happens when the action entropy becomes sufficiently low which indicates is that the agent ends up picking the same bad action and fills up the training samples with bad observations.
The problem is further compounded by the fact that invalid actions do not change the state of the game and we do not do anything to prevent the algorithm from selecting invalid actions, leading to identical training data samples and thus broken learning.

On Fig. \ref{fig:valid moves percentage} and \ref{fig:completion_pct_article} it can be seen that while the agent generally picks valid actions and completes the levels more often than the random agent, it does not reach human-like performance on both seen and unseen levels after level 2.



\subsection{Generalisation}
\label{sec:generalisation}

The observation that the agent does not reach human level performance and sometimes also get stuck on an invalid move may indicate that the agent does not explore sufficiently.
One way to increase exploration with a PPO algorithm is to increase the entropy coefficient which adds an entropy bonus to the loss function ($c_2$ in Eq. \eqref{eg:loss function all}).
Three different configurations were tested: 0.0, 0.001 and 0.01, where 0.001 is the default value.

Generally the learning curves are very similar but the higher the entropy coefficient is, the longer the agent can be trained for and the less likely it is to encounter catastrophic learning behaviours.



Adding color shuffling should also help the agents generalise because it adds randomness.
Indeed, the completion percentage for the CS, EC:0.01 model on level 3 and 4 is better than any of the baseline models and comparable on the other levels.
While it should be noted that the model had been training for longer, this was made possible because of the higher entropy coeifficient and more environment stochasticity.
Color shuffling therefore seems to be a viable strategy in addition to a high entropy coefficient.


\subsection{Max Episode Length}
\label{sec:max episode length}

Using strategies that add randomness and increase exploration are not enough to prevent the agent from sampling the same move over and over again as evidenced by the previous experiments.
We therefore try the strategy of resetting the environment to break the loop if a bad learning behaviour happens.

It should be noted that it is difficult to compare the learning curves of agents trained with reset and those without, since resetting ensures that it is not possible to accumulate large negative rewards from choosing the same invalid action over and over again.
However, the learning curves are still useful for verifying that the agent is improving and not encountering catastrophic learning behaviour.

Using the reset strategy has a large positive impact on the learning.
None of the agents that employ this strategy run into the same loop of selecting the same action all the time which enables the agent to train longer and learn more, with the exception of the CSResetMask model which will be described in the next section.
Resetting the environment after a number of steps is therefore a good strategy that leads to more stable learning.

\subsection{Action Masks}
\label{sec:action mask}

Since none of the other experiments directly prevent invalid actions to be taken, the agent has to first learn to infer which moves that are valid. 
We therefore test two different ways of adding this information -- a hard mask and a soft mask, dubbed V2, as described in Section \ref{sec:methods: action masking}.

Using a hard action mask very quickly leads to high rewards which makes sense since invalid actions lead to a $-0.5$ penalty but are never taken now.
However, as far as stability goes, the training completely fails after around 1.5 million steps, as seen in the sharp drop in the learning curve on Fig. \ref{fig:learning curves}.

Unlike what was seen in many of the previous experiments when the entropy becomes very low/zero, it now receives undefined rewards, indicating something with the algorithm itself is failing.
What is happening is that the action probability distribution from the policy is 100\% of an invalid action, and 0 on the rest, but because of the hard action mask, the final logits distribution is filled with $-\infty$.
Taking the maximum of this vector then leads to unexpected behaviour.
This is supported by the fact that the trained agent is actually not very competent (even worse than random, Fig. \ref{fig:valid moves percentage}) and thus tend to select invalid actions first.

The picture is completely different when using it as a soft action mask.
Looking at Fig. \ref{fig:learning curves}, the CSResetMaskV2 agent is both stable during training and learns faster compared to the CSReset agent (i.e. they reach the same learning plateau after 0.5M and 4M steps, respectively).
It also has better completion rate and competency on both test and training levels than any of the other approaches.



\section{Discussion}

\begin{figure*}[ht]
    \includegraphics[width=0.65\columnwidth]{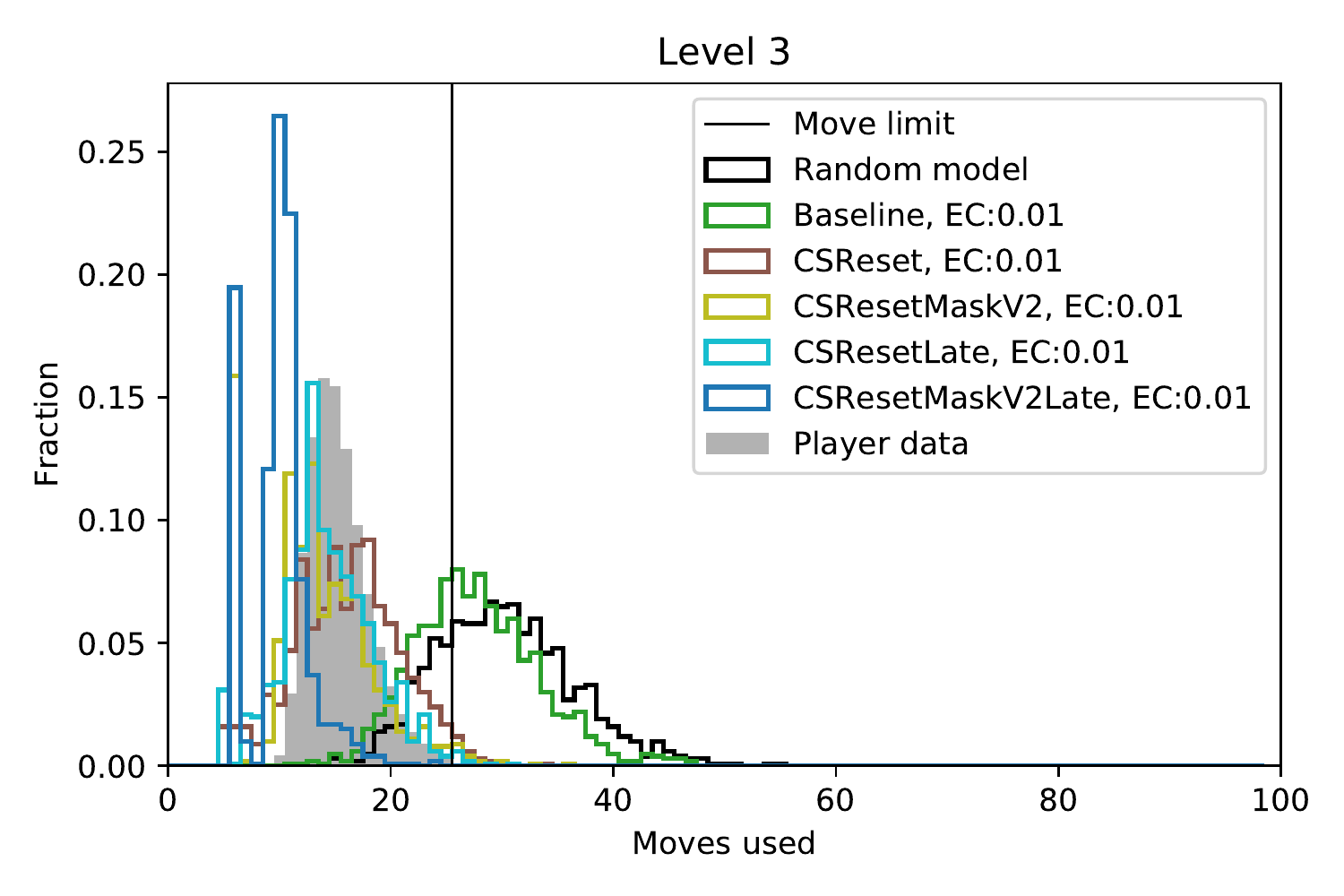}
    \includegraphics[width=0.65\columnwidth]{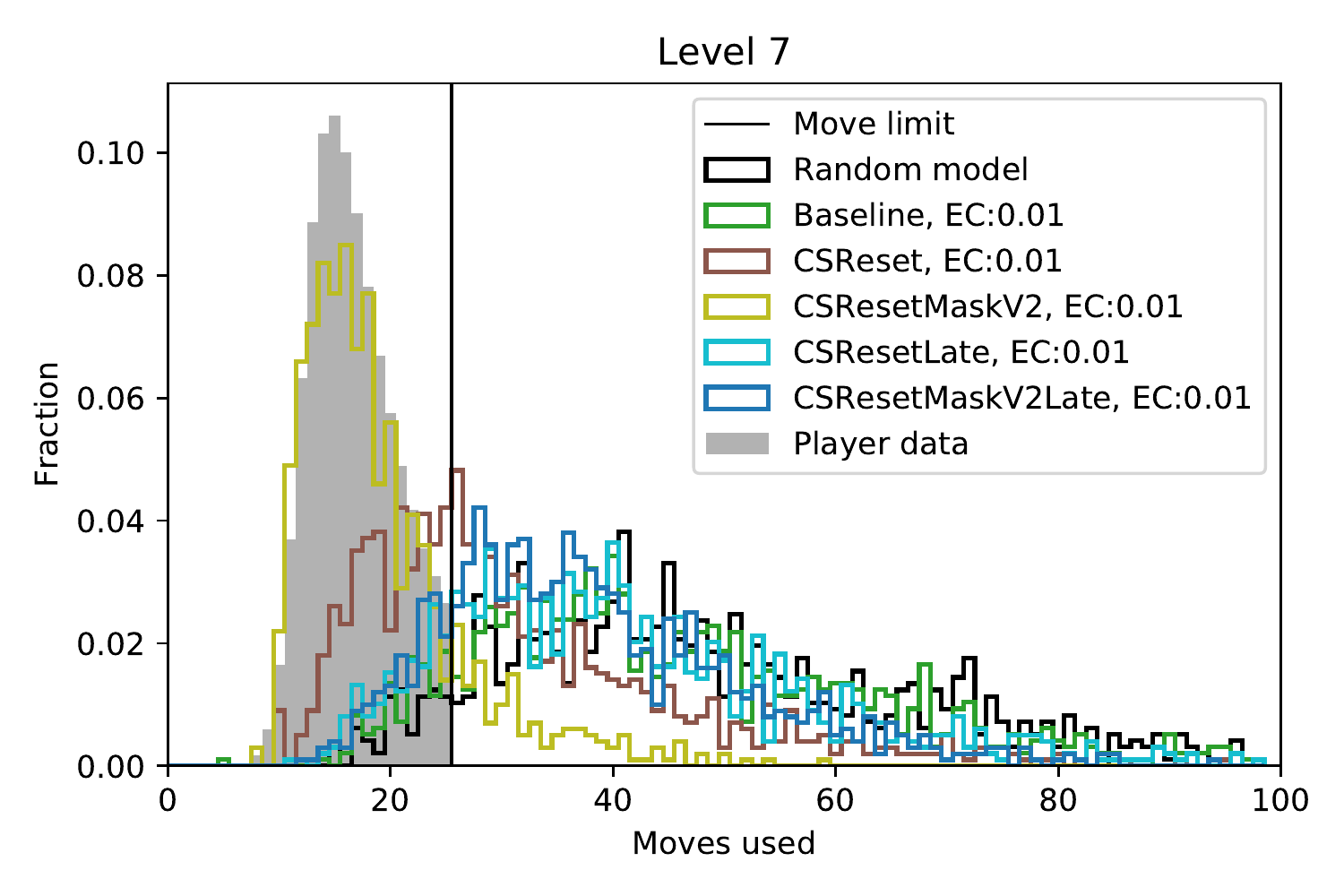}
    \includegraphics[width=0.65\columnwidth]{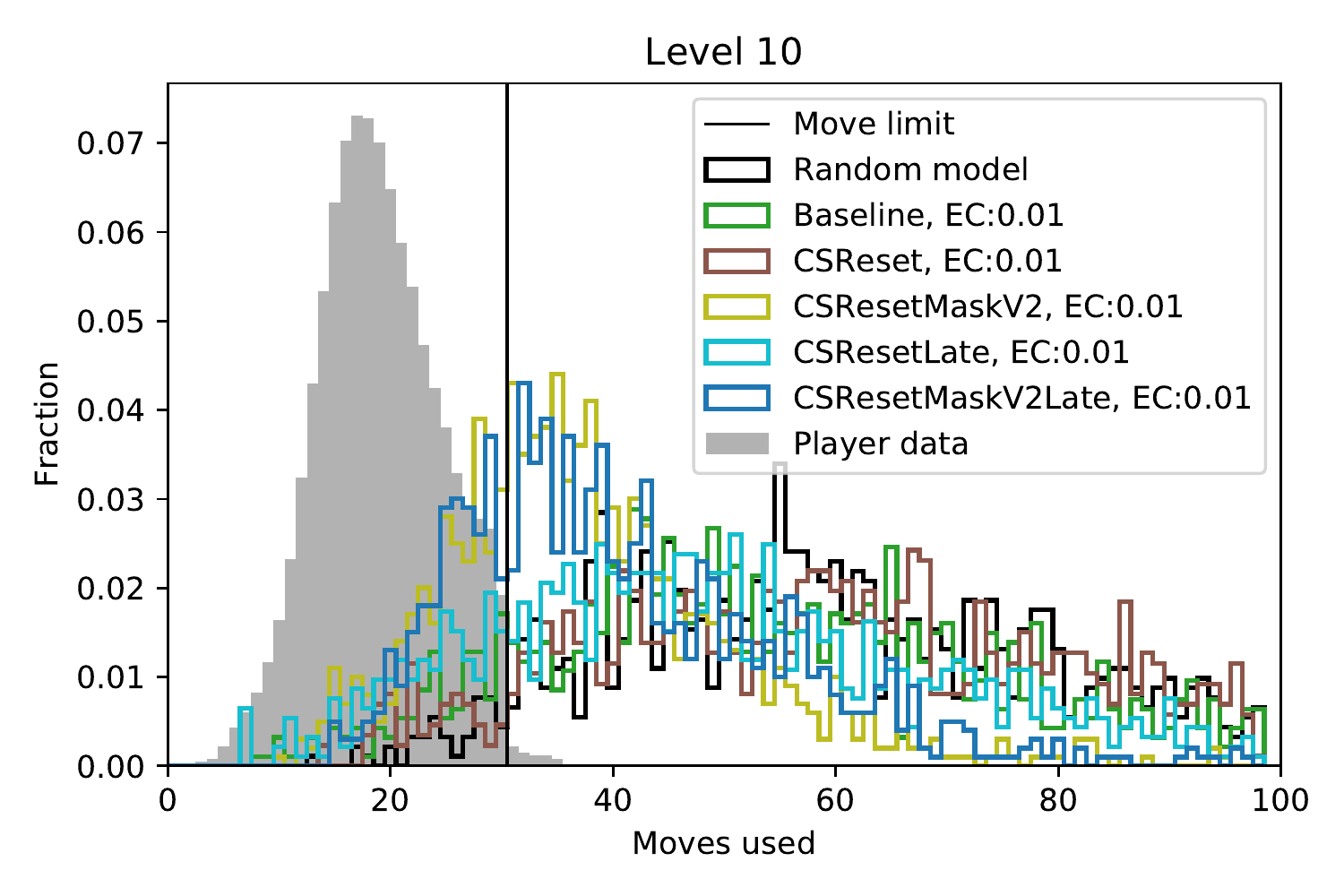}
    \caption{Histograms of how many moves required to finish selected levels for selected agents, including the random agent in black, that show super-, normal and sub-human performance. The grey shaded area is actual player data and is the distribution that we want to mimic. A sharp cut-off can be seen in the player data distributions which is aligned with the move limit. The reason a small tail can be seen in level 10 is because players are able to purchase an additional 5 moves if they fail, but only a fraction of players choose to do so. The normalisation is therefore also not completely comparable since the agents are allowed to play past the move limit.}
    \label{fig:movesused histograms}
\end{figure*}

The most effective strategy for training seems to be resetting the environment after a number of total steps.
Color shuffling together with an increased entropy coefficient are also strategies that help the agent learn despite slowing down the training.
Shifting towards more exploration and less exploitation in games like Lily's Garden therefore seems to be beneficial.

Some of the strategies did not work very well, though, like using a hard action mask or training for too long.
This gives rise to some concerns if used in a production environment and will be discussed below.


\subsection{Dealing With Invalid Actions}

The main issue encountered throughout the experiments was invalid actions, which may be very specific to our environment and implementation of PPO.
For example, in ML-Agents a small probability $\epsilon$ is added to the raw probabilities ensuring that there will be no $-\infty$ when converting to logits.
This avoids the hard action mask problem, but it can be argued that it is not a hard action mask anymore.
Other ways to deal with sampling the same action over and over could be to use an epsilon-greedy approach or by sampling the way we did it in post-training evaluation but this significantly slows down the training.

While this problem with invalid actions may be a very specific problem to our environment, it still highlights a possible issue that may arise in other similar games where some actions do not progress the game.
Additionally, if a hard action mask is being used, the algorithm runs the risk of masking out every action, leading to unexpected behaviour.
This is an issue since in a number of research papers on play-testing agents it is not clear how the action masking is actually being done even though it has a huge impact on the training of the agent.
This adds further complexity to understanding the algorithms and reflects the thoughts in \cite{Ilyas2018} that the implementation matters.

One question is whether using the action mask is practical in the long run since it does require some kind of modelling of the environment.
Additionally, while the levels considered in this paper do not have very complex game mechanics, later levels include mechanics that prevent certain actions.
While the environment could be configured to return a proper action, this may prove computationally and developer-time intensive and therefore not viable in the long run.
However, interestingly enough it was found out during the evaluation process that the action mask in some very specific cases allowed invalid actions.
Whether that is because a bug in the game or action mask modelling is not clear, but it was interesting that this was not a problem when using the soft action mask.
This suggest that using even a imperfect forward model of the environment still improves learning.

\subsection{Usefulness in Production}

There are two things to consider before judging if the agent is actually useful to level designers.

The first question is whether the level designers would be able to rely on the agent or not.
For that to be the case, the more consistent and performant it is, especially on unseen levels, the better.
However, what is observed is that the completion rate is worse on unseen levels.
This limits the usefulness to level designers since the new levels will obviously not have been encountered before.
One solution for this could be to allow the agent to train on the unseen levels.
To see whether this is feasible in a production setting requires further testing.

One other thing to take note of is the fact that the completion rate is low despite picking valid action most of the time.
This suggests that the agents learn how to play the game but not how to play it optimally.
This may be a consequence of the reward function, though -- a relatively big penalty is given for selecting invalid moves compared to collecting objectives.
The first thing the agents learn is therefore how to not take an invalid action.
Learning new things, such as going after objectives, is secondary and would require more training without overfitting.
The best way to achieve this would be to introduce more levels, which should help with generalising and making the agent more consistent.

The second thing to consider is that it must play like a human and not superhuman, so the estimated difficulty matches with how players perceive it.
While the completion percentage used in the post-training evaluation already reflects this aspect, it does not tell the whole story.
Another way to judge how human-like the agent behaves is by looking at the distribution of moves required to finish the level (Fig. \ref{fig:movesused histograms}) and comparing with human data.
This kind of visualisation is also more useful to level designers since it can be used to determine the move limit.
However, none of the models are consistent in being super/sub-human which must be addressed first.

\subsection{Future Work}

When an agent first tries to learn how to play these puzzle games, it first needs to figure out how to do a valid move.
As revealed by using an action mask, it learns much faster if something guides it initially.
One way could therefore be to use imitation learning to first teach it how to do the basics. 
This also has the added benefit that it may be easier to guide the agent to play more like a human which would make the tool more useful to level designers.
It would require some time and effort to set this up in practice, though, both in regards to implementing it in production code but also the time level designers would have to spend training the agent.
Evaluating which approach is more time-effective should therefore not only include computation time but also the human resources required.
However, an imitation learning module has been added to ML-Agents and may provide a good starting point.

The post-training evaluations show that the agents play some levels well but struggle with others.
It therefore seems like a better strategy to spend more time training on the difficult levels rather rather than continuing selecting the levels randomly. 
An idea could be an automated approach like in \cite{Graves2017} where the training examples that yield the most learning are chosen.
This would also open up for training on more levels which should help generalisation of the agent on unseen levels.
One thing to keep in mind before training on many new levels and mechanics, though, is that the agent may be prone to catastrophic forgetting \cite{Kirkpatrick2017} where previously learned behaviours are completely forgotten.

\section{Conclusion}

In this research paper we have successfully adapted the popular RL method PPO to a production grade puzzle game for training play-testing agents.
Crucial to this success, not considering hyper-parameter tuning, was introducing a reset strategy where the environment is reset after a fixed number of steps.
This ensured a more stable training, enabling the models to learn more.
Other strategies also improved other aspects of the training -- color shuffling improved generalisability, and introducing an action mask as a partial forward model of the environment in the observation greatly improved training speed, though the latter may not always be feasible in other types of games.

When we experimented with a hard action mask that was added to the logits of the action probabilities, the algorithm completely broke down.
This happened because all the valid actions from the model were practically 0 while the invalid actions were all 0 because of the action mask, effectively masking out every action and leading to unexpected behaviour.
Various RL libraries use a similar method but it should be used with great caution.
A better approach would be to include the action mask in the observations and thus serving as a partial forward model.

\section{Acknowledgements}

This work has been supported by the Innovation Fund Denmark and Tactile Games.

We thank Hunter Park (https://github.com/H-Park) and Kenneth Tang (https://github.com/ChengYen-Tang) for discussions on the various implementations.

We also thank Rasmus Berg Palm (ITU) for helpful comments for the manuscript.

\bibliographystyle{plain}
\bibliography{referencesmanual,referencesfixed}







\end{document}